\documentclass[10pt,twocolumn,letterpaper]{article}

\usepackage[pagenumbers]{cvpr} % To force page numbers, e.g. for an arXiv version

\usepackage{times}
\usepackage{epsfig}
\usepackage{graphicx}
\usepackage{amsmath}
\usepackage{amssymb}
\usepackage{booktabs}
\usepackage{multirow}
\usepackage{float}
\usepackage{multicol}
\usepackage{lipsum} % for dummy text only
\usepackage{xargs}  % Use more than one optional parameter in a new commands
\usepackage{xcolor}  % Coloured text etc.
\usepackage{dsfont}

\usepackage[pagebackref=true,breaklinks=true,colorlinks,bookmarks=false]{hyperref}
\usepackage[numbers,sort,compress]{natbib} % reference in increasing order
\usepackage[capitalize]{cleveref}
\crefname{section}{Sec.}{Secs.}
\Crefname{section}{Section}{Sections}
\Crefname{table}{Tab.}{Tables}
\crefname{table}{Tab.}{Tabs.}
\newcommand*{\newcite}[1]{~\cite{#1}}

\usepackage[boldmath]{numprint}
\npdecimalsign{.}
\nprounddigits{2}
\newcommand{\round}[1][]{\numprint}

\def\eg{e.g.,~}      % for example
\def\ie{i.e.,~}      % that is, in other words
\def\etal{et al.}      % for example
\def\wrt{w.r.t\onedot}
\newcommand{\comment}[1]{}
\newcommand\norm[1]{\left\lVert#1\right\rVert}

\def\cvprPaperID{5267} % *** Enter the CVPR Paper ID here
\def\confName{CVPR}
\def\confYear{2022}

\begin{document}

\title{What Matters For Meta-Learning Vision Regression Tasks?}
% \title{Meta-Learning Regression Tasks on High-Dimensional Input}

\author{Ning Gao$^{1, 2}$ \ \  Hanna Ziesche$^{1}$ \ \  Ngo Anh Vien$^{1}$ \ \  Michael Volpp$^{2}$ \ \  Gerhard Neumann$^{2}$\\
$^{1}$Bosch Center for Artificial Intelligence \ \  $^{2}$Autonomous Learning Robots, KIT\\
{\tt\small \{ning.gao, hanna.ziesche\}@de.bosch.com} \ \ {\tt\small anhvien.ngo@bosch.com}\\
{\tt\small \{michael.volpp, gerhard.neumann\}@kit.edu}
}
\maketitle

%%%%%%%%% ABSTRACT

\begin{abstract}
Meta-learning is widely used in few-shot classification and function regression due to its ability to quickly adapt to unseen tasks. However, it has not yet been well explored on regression tasks with high dimensional inputs such as images. This paper makes two main contributions that help understand this barely explored area. \emph{First}, we design two new types of cross-category level vision regression tasks, namely object discovery and pose estimation of unprecedented complexity in the meta-learning domain for computer vision. To this end, we (i) exhaustively evaluate common meta-learning techniques on these tasks, and (ii) quantitatively analyze the effect of various deep learning techniques commonly used in recent meta-learning algorithms in order to strengthen the generalization capability: data augmentation, domain randomization, task augmentation and meta-regularization. Finally, we (iii) provide some insights and practical recommendations for training meta-learning algorithms on vision regression tasks. \emph{Second}, we propose the addition of functional contrastive learning (FCL) over the task representations in Conditional Neural Processes (CNPs) and train in an end-to-end fashion. The experimental results show that the results of prior work are misleading as a consequence of a poor choice of the loss function as well as too small meta-training sets. Specifically, we find that CNPs outperform MAML on most tasks without fine-tuning. Furthermore, we observe that naive task augmentation without a tailored design results in underfitting. 

\end{abstract}

%%%%%%%%% BODY TEXT
\begin{figure}[t]
    \includegraphics[width=0.95\columnwidth]{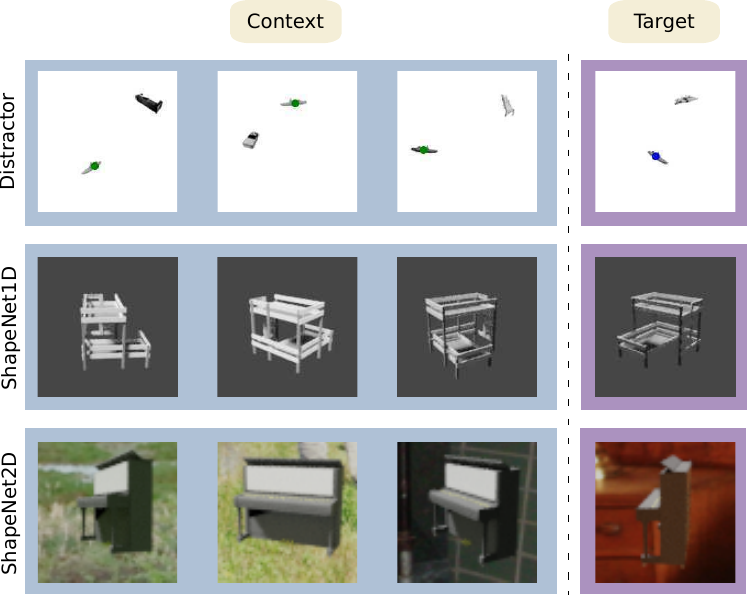}
    \caption{Meta-learning vision regression tasks are designed to i) identify the queried object from context and predict its position for target images (Distractor),
    % from context images $x_c$ and the positions $y_c$ of the queried object (watercraft in this example), the model is capable to identify the features of queried object and predict the location $y_t$ of arbitrary new queried images $x_t$. 
    ii) identify the object's canonical pose from context % (images $x_c$ and corresponding rotation $y_c$ in azimuth axis) 
    and predict the 1D rotation relative to the canonical pose for target images (ShapeNet1D), iii) predict the 2D rotation \wrt the canonical pose with random background (ShapeNet2D). Predictions are performed on unseen objects.}\label{fig:overview}
\end{figure}
\section{Introduction}

Humans are able to rapidly learn the fundamentals of new tasks within minutes of experience based on prior knowledge. For instance, humans can classify novel objects by capturing the distinguishable properties (\eg textures, shapes and scales) from only a few examples.
% \comment{where the discriminative ability is associated with prior knowledge}
% \comment{In many real-world applications, it is preferable to acquire relative small but related datasets rather than single but large dataset in order to learn versatile and transferable skills. }% For instance, the robot learning to push the drawer is also expected to be able to pull the drawer back. Furthermore, the robot should also adapt quickly to new but similar tasks such as pushing the door without extensive data. 
Meta-learning is proposed to learn relevant knowledge from various tasks and generalize to unseen tasks with only a few samples. Of the various meta-learning algorithms, MAML-based models\newcite{MAML, reptile, online-maml, BayesianMAML} and Neural Processes (NPs)\newcite{CNP, NP, ANP, CCNP} are two variants which are receiving increasing attention in the recent years. Both algorithms try to learn good prior knowledge from related tasks without expanding the learned parameters or sacrificing efficiency at inference. While these methods have shown promising results in many domains, such as few-shot classification\newcite{reptile, relation-network, prototype-network, Meta-Semi, matching-network} and hyperparameter optimization\newcite{hyperp-NP, Volpp2020Meta-Learning, bilevel-hpo}, an extensive study on meta-learning vision regression tasks has not yet been conducted. This is in particular true for NPs which have mostly been investigated on tasks with low-dimensional input such as function regression or pixel-wise completion\newcite{DSVNP, NPODE, FNP, RNP}. 

In this paper, we make two major contributions to the largely unexplored area of meta-learning on high-dimensional input tasks. On the algorithmic level, inspired by SimCLR\newcite{simclr}, we propose an improvement to NPs by employing contrastive learning at the functional space (FCL) and still train the model in an end-to-end fashion. On the experimental side, we propose two application datasets, object discovery and pose estimation, which are based on high-dimensional inputs and require the meta-learning models to learn and reason at an image level. 
% \comment{These two applications serve us as a testbed, in order to }Afterwards, we investigate different algorithmic choices as well as data and task augmentation techniques for meta-learning on these two applications.
% \comment{In order to evaluate their performance on high-dimensional inputs, \checklater{we consider two applications which require the model to learn and reason at an image level, namely object discovery and pose estimation.}} 

For the first application we create a regression task called ``Distractor" (see \cref{fig:overview}), where each image contains two objects, the queried object and a distractor object, placed at random positions. The goal of this task is to identify the queried object and predict its position in the image plane. Unlike previous tasks such as image completion, where each pixel is considered as an independent input, our task requires the model to learn a high-level representation from the entire image.
% we create an image-based regression task called Distractor. Being different from the image completion problem which considers each pixel as independent input, Distractor contains a queried object and a distractor in each image placed at random positions. The idea is to identify and predict the location of the queried object, which essentially requires the model to capture and distinguish features of the queried object from the distractor. For example, in \cref{fig:overview} the model needs to first identify the watercraft as the queried object from a context set by predicting a meaningful task representation which is distinguishable from the distractors. Afterwards, the model is expected to predict the location of a queried object conditioned on the predicted task representation given a different distractor. 
The second application (\ie pose estimation) is inspired by prior work\newcite{MWM, MRM, TaskAugmentation, data-augmentation-meta-learning} on the Pascal1D dataset. As this dataset shows limited object variations and features only 1D rotation around the azimuth axis, we generate two new datasets with increasing task diversity, \eg by introducing random background, cross-categorical object variations and 2D rotation. Since the background is generated from real-world images instead of blank as in prior work, our datasets significantly increase the task difficulty and allow us to perform a thorough investigation of the performance for the considered meta-learning approaches. Examples of our datasets are shown in \cref{fig:overview} where i) ShapeNet1D contains 1D rotations as in Pascal1D, however with larger object variations and ii) ShapeNet2D features 2D rotation and random background.

For both applications, we evaluate the performance on novel objects at both {\bf intra-category} (IC) and {\bf cross-category} (CC) levels. The results on Distractor show that our proposed algorithmic improvements significantly increase the performance, indicating our methods can enhance the task expressivity. The results on pose estimation demonstrate that meta-learning can successfully be applied to predict poses of unknown objects, which has a huge potential in robotic grasping and virtual/augmented reality (VR/AR).

Prior work\newcite{MWM, MRM, TaskAugmentation, data-augmentation-meta-learning} on Pascal1D also demonstrates that meta-learning algorithms suffer from overfitting, especially with limited training data. Our work analyzes the effect of different techniques commonly adopted in recent meta-learning methods (\ie data augmentation, task augmentation, regularization and domain randomization) on aforementioned datasets. We empirically find that the meta-learning algorithms employed in our work ultimately lead to overfitting regardless of dataset size for both applications. Moreover, our work shows that the results in prior work\newcite{MRM, MWM}, where MAML typically performs best for such tasks, are misleading. In particular, we find Conditional Neural Processes (CNPs)\newcite{CNP} are more flexible and efficient than MAML in the investigated pose regression tasks. Additionally, we find that MAML\newcite{MAML} suffers from underfitting especially on large-scale datasets and depends heavily on hyperparameter tuning. 
% \comment{Moreover, we find that using cross-attention (CA) can alleviate the problem of overfitting on pose estimation but not on the Distractor task, which indicates that cross-attention is more effective on object-centric inputs. }
% More results are shown in \cref{experiments}.
% without sacrificing computational time using fast attention module from Performer\newcite{Performers}. 

The primary contributions of this work can be summarized as follows: (1) We investigate meta-learning algorithms on vision regression tasks and demonstrate their ability to tackle structured problems. (2) We propose functional contrastive learning on the task representation of CNPs and thereby improve its expressivity. (3) We quantitatively analyze various deep learning techniques to alleviate meta overfitting. Our results rectify misleading conceptions from prior work, e.g., that MAML performs best for such tasks. We also present insights and practical recommendations on designing and implementing meta-learning algorithms on vision regression tasks.
% and hope our work facilitates future work on meta-learning vision regression tasks. 

\section{Related Work}

% \paragraph{Meta-Learning.}
\noindent\textbf{Meta-Learning.}
In meta-learning, also known as \textit{learning to learn}, a learning agent gains meta knowledge from previous learning episodes or different domains and then uses this acquired knowledge to improve the learning on future tasks\newcite{metalearning-review}. MAML is an optimization-based meta-learning method and represents the meta knowledge as the model parameters, where learning good initial parameters can enable quick adaptation to new tasks with only few update steps on a small number of samples\newcite{MAML}. Different from MAML, Neural Processes (NPs) constitute a class of neural latent variable models and interpret meta-learning as conditional few-shot function regression\newcite{NP}. Similar to Gaussian Processes, NPs model distributions over functions conditioned on contexts\newcite{NP, ANP, VERSA}. Meta-learning algorithms have been applied successfully in low-dimensional function regression\newcite{CNP, NP, ANP, DSVNP}, image completion\newcite{CCNP, FNP, RNP}, few-shot classification\newcite{reptile, relation-network, prototype-network, Meta-Semi, matching-network, MMAML}, reinforcement learning\newcite{BayesianMAML, MetaRL-offline, Metaworld, MetaRL-structure, contrastiveMLRepresentation}, and neural architecture search (NAS)\newcite{DARTS, Zoph2017NeuralAS, Elsken2020MetaLearningON, Lee2021RapidNA}. Recent works\newcite{online-maml, MRM, MWM, TaskAugmentation} go one step further and apply meta-learning to pose estimation using gray-scale images. However, in these studies, the prediction is restricted to 1D rotation and the employed loss function is ill-posed as it does not take the periodicity of rotation into consideration. Moreover, \newcite{FRCL} proposes to improve meta-learning by adding contrastive representation learning from disjoint context sets. A follow-up work\newcite{CLNP} further extends this idea to time series data by combining contrastive learning with ConvNP\newcite{CCNP}. However, in contrast to these two methods which need to learn a representation in a self-supervised way and fine-tune on downstream tasks subsequently, we use functional contrastive learning (FCL) between context and target sets and train in an end-to-end fashion. 
% In our work, we investigate 2D pose regression on RGB-images with additional random background of real images, where the increasing task difficulty further presents how expressive the latent variable models can achieve. Furthermore, we propose a new paradigm of vision-based regression task called Distractor, which arises more interest in the applications of exploring meta-learning algorithms.
% \subsection{Add Contrastive Learning?}
\\

% \paragraph{Meta Overfitting.}
\noindent\textbf{Meta Overfitting.}
It is well-known that meta-learning algorithms suffer from two notorious types of overfitting: i) \textbf{Memorization overfitting} occurs when the model only conditions on the input to predict the output instead of relying on the context set\newcite{MWM}; ii) \textbf{Learner overfitting} happens when the prediction model and meta-learner overfit only to the training tasks and cannot generalize to novel tasks even though the prediction can condition on the context set\newcite{MRM}. Recently, different methods have been proposed to mitigate those overfitting issues, \eg adding a regularization term on weights to restrict the memorization\newcite{MWM}. However, tuning a regularization term between underfitting and overfitting is challenging\newcite{regularization}. Subsequently, a related work\newcite{MRM} applied task augmentation which helps both memorization and learner overfitting. Meanwhile, \newcite{TaskAugmentation} proposed MetaMix and Channel Shuffle to linearly combine features of context and target sets and replace channels with samples from different tasks. Furthermore, Ni \etal\newcite{data-augmentation-meta-learning} empirically showed that data augmentation can also alleviate meta overfitting. Moreover, they find that employing data augmentation on target set achieves better performance. However, extensive comparisons on how these methods perform individually or combinedly are missing. In this work, we separate these techniques into data augmentation (DA), task augmentation (TA), meta-regularization (MR) and domain randomization (DR), and quantitatively compare them in different combinations on the two aforementioned applications in order to arrive at a better understanding and consistent comparisons.

%MAML aims to learn a good prior which could benefits all the sampled tasks and assumes the optimal parameters could be reached within few steps updates.

% \input{docs/6_background}
% \input{docs/3_approach}
\section{Study Design}
\label{sec:task_description}

%\subsection{Notations}
\label{sec:background and notations}
We now briefly describe both MAML and CNP in a unified way. We assume that all tasks are sampled from the same distribution $p(\mathcal{T})$, each task $\mathcal{T}_i$ includes a context set $\mathcal{D}_C^i=\{(x_{C, 1}, y_{C, 1}),...,(x_{C, K}, y_{C, K})\}_i$ and a target set $\mathcal{D}_T^i=\{(x_{T, 1}, y_{T, 1}),...,(x_{T, M}, y_{T, M})\}_i$ where $K$ and $M$ are the number of samples in each set which could be different for each task. The entire training dataset is denoted as $\mathcal{D} = \{\mathcal{D}_C^i, \mathcal{D}_T^i\}_{i=1}^N$ where N is the number of tasks sampled for training. During inference, the model is tested on a new task $\mathcal{T}^*\sim p(\mathcal{T})$ given a small context set, from which it has to infer a new function $f^*: (\mathcal{D}_C^*, x_T^*) \rightarrow \hat{y}_T^*$. In meta-learning, there are two types of learned parameters, the first is the meta-parameters $\theta$, which are learned during a meta-training phase using $\mathcal D$. The second is task-specific parameters $\phi^*$ which are updated based on samples from a new task $\mathcal{D}_C^*$ conditioned on the learned meta-parameters $\theta$. Predictions can be constructed as $\hat{y}_T^*=f_{\theta, \phi^*}(x_T^*)$, where $f$ is the meta-model parameterized by $\theta$ and $\phi^*$.
% Both MAML and NPs form task-specific $\phi^*$ as $p(\phi^*|D^{*}_{c}, \theta)$ which conditioned on meta-parameters $\theta$ and context set sampled from task. However, in 

MAML considers both $\theta$ and $\phi^*$ as weights of neural networks 
% and $\phi$ is updated by gradient optimization on new task. 
while CNP considers only $\theta$ as neural weights. Different from MAML, which updates $\phi^*$ by gradient optimization on the new task samples, CNP takes $\phi^*$ as task representation and predicts it from the context set as $\phi^*=\bigoplus_{i=1}^Kh_{\theta}(x_{C, i}^*, y_{C, i}^*)$. Here $\bigoplus$ is a permutation invariant operator% since task representation should not be affected by context orders
, $h$ is an encoder parameterized by $\theta$. Subsequently, a decoder $g_\theta$ will take $\phi^*$ as an additional input and output $\hat{y}_T^*=g_{\theta}(x_T^*, \phi^*)$. 
% Note that Ctxs $\theta=\{\gamma, \psi\}$ are fixed after meta-training phase, therefore CNP doesn't not require any fine-tuning as MAML.
Note that meta-parameters $\theta$ are fixed after meta-training phase, therefore CNPs don't not require any fine-tuning as MAML.

% \begin{figure*}
% 	\centering
% 	\includegraphics[scale=0.25]{figs/airplane_distractor.png}
% 	\caption{An example of generated images for object identification and localization. In this example, the target object ``car" is randomly located together with distractor which sampled from different categories. This task aims to enable the model to first identify the target object and predict the location under the interference of distractor.}
% 	\label{fig:distractor}
% \end{figure*}

\subsection{Problem Setting}
\label{problem_setting}
% Considering regression tasks with high dimensional input using neural process models is not well explored, 
In this paper, we consider two types of image-based regression tasks, namely object discovery and pose estimation. First, we propose a non-trivial object discovery task called Distractor, which is only used for evaluating CNP variants. In contrast to existing object detection tasks\newcite{MSCOCO, KITTI, PascalVOC, ILSVRC} that are designed to specify all object instances from an input image, our task aims to i) distinguish the queried object from other distractors and additionally ii) predict its 2D location in the image plane. Therefore, it is essential to learn a distinctive embedding $\phi^*$ that can represent various queried objects given their associated context images $\{x_{C,i}^*\}_{i=1}^K$ and corresponding positions $\{y_{C,i}^*\}_{i=1}^K$. Note that the distractors are sampled randomly from all categories and in many cases their appearances closely resemble the queried object. Hence, it is expected that aggregating multiple context pairs helps extracting expressive information to disambiguate the tasks and thus improve the performance.
% Hence, aggregating multiple context pairs is expected to fuse more expressive information in order to disambiguate the tasks and improve the performance.

The second task, pose estimation, is evaluated on three datasets, namely Pascal1D, ShapeNet1D and ShapeNet2D with incremental difficulty, caused \eg by extending inference to unseen cross-category objects, adding random backgrounds and extending 1D rotations to 2D rotations. Note that in this task, each object has a random canonical pose, which has to be learned from a context set $D_C^*$ where $\{y_{C,i}^*\}_{i=1}^K$ are the ground-truth rotations of context images $\{x_{C,i}^*\}_{i=1}^K$. 

We use these tasks for an exhaustive evaluation of meta-learning algorithms: i) We evaluate the performance of CNPs using different aggregation operators, \ie mean\newcite{CNP}, max, bayesian aggregation (BA)\newcite{BayesianNP} and cross-attention (CA)\newcite{ANP}. ii) We evaluate MAML on Pascal1D and ShapeNet1D following \newcite{MRM, MWM} and compare it with different CNP variants. iii) Furthermore, we investigate meta overfitting with respect to different choices, \eg augmentations, regularization, aggregation operators and task properties. iv) Moreover, we combine functional contrastive learning (FCL) with CNPs and compare it with original CNPs. 
%where Pascal1D is proposed and conducted by prior work\newcite{MWM, MRM, TaskAugmentation}, 
%including 65 objects in total where each object contains 100 images with random 1D rotation in azimuth. However, this dataset has limited task diversity and uses inappropriate objective function in prior work which will be discussed in detail in \cref{taskdescription:Pose Estimation}. 
%Therefore, we create ShapeNet1D and ShapeNet2D datasets. ShapeNet1D is similar as Pascal1D but with larger task diversity and modified objective function while ShapeNet2D is generated with 2D rotation and using random background instead of static blank scene.

% \subsection{contrastive learning neural processes}

% \subsection{Entropy loss}

\subsection{Datasets}
% Most current famous object detection tasks, for instance, MSCOCO\newcite{MSCOCO}, KITTI\newcite{KITTI}, Pascal VOC\newcite{PascalVOC} and ILSVRC\newcite{ILSVRC}, encourage methods to detect and classify all objects in the image. We create a object discovery task, in contrast, aims to identify and predict solely the properties of the queried object insteaf of others. More spacifically, each image contains two objects with random position. The model is essential to learn which object is queried from context set and further predict its 2D position in image plane. Note that this task is non-trivial since traditional object detector cannot differentiate the queried object. 
% \update{move to experiment section: CNP\newcite{CNP} can be interpreted as a conditional model, where each object is encoded as a task and the prediction is conditioned on task representation. We evaluate CNP on Distractor with respect to different aggregation methods instead of only using mean aggregation in the original paper}. 
We generate \textbf{Distractor} that contains 12 object categories from ShapeNetCoreV2\newcite{ShapeNet}, where each category includes 1000 randomly sampled objects. For each object we create 36 $128\times128$ gray-scale images, containing two objects with random azimuth rotation and 2D position (see \cref{fig:overview}). The data generation is based on an extended version of a prior open-source pipeline\newcite{VERSA}. We choose 10 categories for training, where we reserve $20\%$ of the data for intra-category (IC) evaluation. The remaining 2 categories are only used for cross-category (CC) evaluation. The second dataset, \textbf{Pascal1D}\newcite{MWM}, contains 65 objects from 10 categories. We randomly select 50 objects for training and the other 15 objects for testing. 100 $128\times128$ gray-scale images are rendered for each object with a random rotation in azimuth angle normalized between $[0, 10]$. Since the performance is limited due to the size of the dataset, we generate a larger dataset, \textbf{ShapeNet1D} which includes 30 categories. 27 of these are used during training and IC evaluation, the other 3 categories are used for CC evaluation. For each training category, we randomly sample 50 objects for training and 10 for IC evaluation while CC evaluation is performed on 20 objects for each unseen category. To further increase the task difficulty, we create \textbf{ShapeNet2D} which includes 2D rotations. We restrict the azimuth angles to the range $[0^\circ, 180^\circ]$ in order to reduce the effect of symmetric ambiguity while elevations are restricted to $[0^\circ, 30^\circ]$. Furthermore, we use RGB images and employ randomly sampled real-world images from SUN2012\newcite{SUN2012} as background instead of static background. 
% Note that all tasks are evaluated on both intra- and cross-category level except Pascal1D.

\subsection{Data Augmentation, Domain Randomization, Task Augmentation and Meta Regularization}
\label{exp:augmentation}

\textbf{Data Augmentation (DA).} We use standard image augmentation tecnniques in our work, \ie \textit{Dropout} and \textit{Affine} for all tasks, and an additional \textit{CropAndPad} for all pose regression tasks. Furthermore, we employ \textit{Contrast}, \textit{Brightness} and \textit{Blur} for ShapeNet2D. Details are presented in \cref{app:data_augmentation}. 

\textbf{Domain Randomization (DR).} For ShapeNet2D, we additionally employ DR\newcite{DR-sim2real} by regenerating background images for all training data after every 2k training iterations while the data used for evaluation remain the same. 

\textbf{Task Augmentation (TA).} Task augmentation adds randomness to each task in order to encourage the meta-learner to learn non-trivial solutions instead of simply memorizing the training tasks. Following \newcite{MRM}, we sample random noise $\epsilon^{(t)}$ from a discrete set for each task and create new tasks by adding the noise to the regression targets: $D_C^{(t)}=\{x_{C, i}^{(t)}, \ y_{C,i}^{(t)}+\epsilon^{(t)}\}_{i=1}^K$ and $D_T^{(t)}=\{x_{T, i}^{(t)}, \ y_{T,i}^{(t)}+\epsilon^{(t)}\}_{i=1}^M$. Specifically, we sample 2D position noise from a discrete set $\epsilon\in\{0, 1, 2,..., 16\}^2$ for Distractor. For Pascal1D, we use the same noise set $\{0., 0.25, 0.5, 0.75\}$ as proposed in \newcite{MWM, MRM} while $\{0., 0.125, 0.25,..., 2\}$ for ShapeNet1D. In ShapeNet2D, we first only add random noise in the azimuth angle from the discrete set $\{-10^{\circ}, -9^{\circ}, ..., 20^{\circ}\}$ and in a second step add additional elevation noise from the set $\{-5^{\circ}, -4^{\circ}, ..., 10^{\circ}\}$ for further comparison.

\textbf{Meta Regularization (MR).} Following Yin et al.\newcite{MWM}, we employ MR on the weights $\theta$ of the neural networks. Furthermore, we find that it is crucial to fine-tune the coefficient $\beta$ which modulates the regularizer and task information stored in the meta-parameters $\theta$. In our experiments, we use $\beta=1e^{-4}$ for Pascal1D, $1e^{-7}$ for ShapeNet1D and ShapeNet2D. More details about MR are presented in \cref{app:meta-regularization}.

\subsection{Functional Contrastive Learning (FCL)}
% We compare the performance using different aggregations, namely mean\newcite{CNP}, Max, BA\newcite{BayesianNP} and cross-attention\newcite{ANP}. Moreover, 
The representations learned by CNP are invariant under permutation of the elements within a given context set. This property is achieved by a permutation invariant aggregation mechanism, \eg max aggregation. However, another desirable property of the representation is invariance across context sets of the same task. 
% \comment{CNP considers that task representation should be invariant to the permutation of elements in the context set.}
%which lacks consideration on task distribution in the latent space. 
% we consider representation learning of aggregations given different context sets. The learnt 
In particular, the representations of different context sets belonging to the same task should be close to each other in the embedding space, while representations of different tasks should be farther apart.
% \comment{be and repelled between different tasks}. 
To achieve this, we add an additional contrastive loss at the functional space and train the model in an end-to-end fashion. The contrastive cross-entropy loss is defined as follows\newcite{simclr}:
\begin{align}
\label{eq:contrastive_all}
{\mathcal L_{\mathrm{FCL}}} = -\frac{2}{N}\sum_{t=1}^{N}{\log\frac{\exp(\mathrm{sim}(\phi_C^{(t)} \cdot \phi_T^{(t)})/\tau)}{D(\phi_C^{(t)})D(\phi_T^{(t)})}},
\end{align}% {\mathcal L} = \frac{1}{N}\frac{1}{M}\sum_{t=1}^{N}{\sum_{j=1}^{M}{||y_{T,j}^{(t)} - y^{{(t)},*}_{T, j}||_2}} + \frac{1}{N}\sum_{t=1}^{N}{\mathcal L}_c(\phi_C^{(t)}, \phi_T^{(t)}),
% \end{align}
where $N$ denotes the number of tasks per batch.
% $M$ denotes the number of target images per task, $y_T$ and $y_T^*$ the predicted and ground-truth positions of target set. \checklater{As in \newcite{simclr}, the contrastive cross-entropy loss} of each task ${\mathcal L}_c$ is defined as:
% \begin{align}
% \label{eq:contrastive_part}
% {\mathcal L}_c(\phi_C, \phi_T) = -2\log\frac{\exp(\mathrm{sim}(\phi_C \cdot \phi_T)/\tau)}{D(\phi_C)D(\phi_T)},
% \end{align}
$(\phi_C^{(t)}, \phi_T^{(t)})$ denotes a positive pair of latent representations of a given task obtained from context and target set respectively. More specifically, the pairs are obtained via max aggregation $\phi_C^{(t)} = \max(r_{C, 1}^{(t)}, \ldots, r_{C, K}^{(t)})$ and $\phi_T^{(t)} = \max(r_{T, 1}^{(t)}, \ldots, r_{T, M}^{(t)})$, where $K$ denotes the number of context pairs per task and $M$ the number of target pairs per task. $\max$ returns the element-wise maximum over the latent variables $r_i=h_\theta(x_i,y_i)$ which are output by the encoder network $h_\theta$ for each context pair ${(x_i,y_i)}$. 
$\tau$ is a temperature parameter, \comment{similar to \newcite{simclr}, we consider $\tau$ as a hyperparameter} which is crucial for learning good representations (details are presented in \cref{app:FCL}). $\mathrm{sim}(\cdot)$ is the cosine similarity and ${D(\phi_i^{t})}$ sums the similarity of all positive and negative pairs for $\phi_i^{t}$:
% $D(\phi_i^{(t)})=\sum_{k=1}^{N}\sum_{j\in \{C,T\}}{\mathds{1}_{[\{k\neq t\}\lor \{j\neq i\}]}\exp(\mathrm{sim}(\phi_i^{(t)}\cdot \phi_j^{(k)})/\tau})$, 
\begin{equation}
    D(\phi_i^{t})=\sum_{k=1}^{N}\sum_{j\in \{C,T\}}{\mathds{1}_{[\{k\neq t\}\lor \{j\neq i\}]}\exp(\frac{\mathrm{sim}(\phi_i^{t}\cdot \phi_j^{k})}{\tau}}),
\end{equation}
where $\mathds{1}_{[\{k\neq t\}\lor \{j\neq i\}]}\in \{0, 1\}$ is an indicator evaluating to 1 only if the representations are sampled from different tasks or different sets. The log-value in \cref{eq:contrastive_all} can be interpreted as the weighted importance of the positive pair. Therefore, this loss function encourages the model to obtain large similarity for positive pairs and small for negative pairs. 
% \comment{In the experiment section we will refer to the combination of CNP+max aggregation with FCL described here as (Max$_{\rm FCL}$).} 
% \comment{During inference, only $\phi_C$ will be used as the conditional variable since the labels $y_T$ in the target set is not available.}

%\begin{align}
%\label{eq:zizj}
%z_C = \max(r_{c, 1}, r_{c, 2}, ..r_{c, K}) \notag\\
%z_T = \max(r_{t, 1}, r_{t, 2}, ..r_{t, M})
%\end{align}

% \subsection{Experimental Design}
\subsection{Objective Functions and Evaluation Metrics}
\label{section:objective}

\noindent\textbf{Pascal1D.} 
% Because of the limited training data, meta-learning methods tend to simply memorize the training objects instead of learning from context, which hinders generalization on novel objects during test.
Following prior work\newcite{MWM, MRM, TaskAugmentation} we conduct experiments using the MSE score between predicted and ground-truth azimuth rotation for both training and evaluation. However, this loss function does not take the ambiguity of coterminal angles into account. Hence, it can hamper the training process, \eg predicting $359^\circ$ for a ground-truth angle of $0^\circ$ incurs a higher loss than predicting $180^\circ$. Nevertheless, we follow the same setup to obtain a fair comparison to prior works.\\
% \comment{In addition, \checklater{we take data augmentation into comparison which is also missing in prior works}. }
% Afterwards, we make an improvement on the loss function in ShapeNet1D. 
% Furthermore, we take data augmentation into our comparison. 

% Moreover, we test two variants of CNP, the first one uses mean aggregation which is same as \newcite{MWM, MRM}, the second one is with cross attention module which is similar as \newcite{ANP} but only the deterministic path and without self-attention module. Here we denote the second variant as ANP for simplicity. 

% We find this task has some inappropriate design for both training and evaluation.
% First, meta-learning algorithms require large dataset during meta-training in order to have a good generalize ability. This dataset in contrast contains only 50 objects which is far from enough. The results in table \cref{table:pascal1d} also shows the mean prediction error is around 30 degree after denormalizing the value. Thus, it is unpersuasive to say meta-learning can help predict poses of novel objects even with meta augmentation techniques. Another more critical point we found during implementation is that, 

\noindent\textbf{ShapeNet1D.} Instead of using MSE score, we use the ``cosine-sine-loss" for training and a prediction error defined in terms of the angular degree for evaluation. The loss of a single sample is defined as:
% \begin{align}
% \label{eq:loss_shapenet1d}
% {\mathcal L} = \frac{1}{N}\frac{1}{M}\sum_{t=1}^{N}\sum_{j=1}^{M}&|\cos{(y_{t,j})}-\cos{(y^\ast_{t,j})}|^2 \notag\\
% &+ |\sin{(y_{t,j})}-\sin{(y^\ast_{t,j})}|^2 ,
% \end{align}
\begin{align}
\label{eq:loss_shapenet1d}
{\mathcal L} = |\cos{(y)}-\cos{(y^\ast)}|^2+ |\sin{(y)}-\sin{(y^\ast)}|^2,
\end{align}
where 
% $N$ is the number of objects per minibatch, $M$ is the number of test images per object, 
$y^\ast$ is the ground-truth rotation and $y$ the predicted rotation. The prediction error used for evaluation is defined as follows:
% \begin{align}
% \label{eq:metric_shapenet1d}
% {\mathcal E} = \frac{1}{N}\frac{1}{M}\sum_{t=1}^{N}{\sum_{j=1}^{M}{\min\{\mathcal{E}_{y_{t,j}^{+}, y_{t,j}^\ast}, \mathcal{E}_{y_{t,j}^{-}, y_{t,j}^\ast}, \mathcal{E}_{y_{t,j}, y_{t,j}^\ast}\}}},
% \end{align}
\begin{align}
\label{eq:metric_shapenet1d}
{\mathcal E} = {\min\{\mathcal{E}_{y^{+}, y^\ast}, \mathcal{E}_{y^{-}, y^\ast}, \mathcal{E}_{y, y^\ast}\}},
\end{align}
where
\begin{align*}
\mathcal{E}_{y^{\pm}, y^\ast} = |y\pm360 - y^\ast|, %\notag,
%\mathcal{E}_{y^{-}, y^\ast} &= |y-360 - y^\ast|, \notag\\
\mathcal{E}_{y, y^\ast} = |y - y^\ast|.
\end{align*}

\vspace{0.1cm}
\noindent\textbf{ShapeNet2D.} 
% During training, the model need to implicitly encode all relevant information into the latent space apart from background. 
We represent the 2D rotation as quaternion in both training and evaluation. The loss of a single sample is accordingly defined as follows:

% \begin{align}
% \label{eq:metric_shapenet1d}
% {\mathcal L} = \frac{1}{N}\frac{1}{M}\sum_{t=1}^{N}{\sum_{j=1}^{M}{\min\{\mathcal{E}(q_{t,j}, q^\ast_{t,j}), \mathcal{E}(-q_{t,j}, q^\ast_{t,j}})\},}
% \end{align}
% \begin{align}
% \label{eq:metric_shapenet1d}
% \mathcal{L} = \min\{\mathcal{E}(q_{t,j}, q^\ast_{t,j}), \mathcal{E}(-q_{t,j}, q^\ast_{t,j})\}
% \end{align}
% where
% \begin{align}
% \mathcal{E}(q, q^\ast) &= \left|q^\ast - \frac{q}{\norm{q}} \right|,
% \end{align}
\begin{align}
\label{eq:metric_shapenet1d}
\mathcal{L} = \min\left\{\left|q^\ast - \frac{q}{\norm{q}}\right|, \left|q^\ast - \frac{-q}{\norm{q}}\right|\right\},
\end{align}
% where
% \begin{align}
% \mathcal{E}(q, q^\ast) &= \left|q^\ast - \frac{q}{\norm{q}} \right|,
% \end{align}
where $q^\ast$ denotes the ground-truth unit quaternion and $q$ denotes the predicted quaternion. We empirically find that using this objective function achieves a better performance than constraining the scalar part of $q$ to be positive. We hypothesize that enforcing the scalar constraint breaks the continuity of the rotation representation and therefore hampers training.
%We assume the reason is giving additional constraint \checklater{(\ie enforcing the scalar part)} will \checklater{break the continuity of rotation representation and hamper the optimization with unnecessary constraint.}

% narrow down the \ckechlater{training} optimization to a discontinuous search space and thus harm the advantage of using quaternion as loss function.

\section{Experiments}
\label{experiments}
%\resizebox{\textwidth}{!}{
% Second version of table, with booktabs.
In this section we present experimental results\footnote{Codes and data are available at \url{https://github.com/boschresearch/what-matters-for-meta-learning}}, perform a thorough analysis and provide insights and recommendations. Instead of presenting the results following the task sequence, we structure this section by different algorithmic choices and perform a systematic comparison over all tasks by raising different questions. \cref{app:visualization_examples} provides visualization examples of different tasks. \\
\begin{table}[htb!]
	\centering
	\begin{tabular}{llllll}
		\toprule
		Methods     & Mean                     & Max                & BA                  &CA                & Max$_{\rm FCL}$\\
		\midrule
		No Aug      & \round{6.0155}          & \round{5.1122}      & \round{4.6286}      &\round{5.1284}      &\textbf{\round{3.7030}} \\
		            & \round{6.8910}          & \round{6.1730}      & \round{5.9117}      &\round{6.3864}      &\textbf{\round{4.6095}} \\
		DA          & \round{2.6705}          & \round{2.4493}      & \round{2.4422}      &\round{2.6540}      &\textbf{\round{2.0039}} \\
		            & \round{4.0966}          & \round{3.7517}      & \round{3.9666}      &\round{4.0792}      &\textbf{\round{3.0462}} \\
		TA          & \round{6.2860}          & \round{6.1847}      & \round{6.3349}      &\round{6.3208}      &\textbf{\round{5.4466}} \\
		            & \round{7.1862}          & \round{7.0377}      & \round{7.0160}      &\round{7.0211}      &\textbf{\round{6.6585}} \\
	    TA+DA       & \round{3.2020}          & \round{3.0937}      & \round{2.6454}      &\round{3.0526}      &\textbf{\round{2.6011}} \\
		            & \round{6.0665}          & \round{5.1410}      & \round{4.6730}      &\round{4.9838}      &\textbf{\round{3.8993}} \\
		\bottomrule
	\end{tabular}
	\caption{Prediction error (pixel) on euclidean distance in the 2D image plane for Distractor. Different aggregation methods and augmentations are employed\comment{ while Max aggregation is further combined with FCL (Max$_{\rm FCL}$)}. The first row shows results for intra-category (IC) evaluations, the second row for cross-category (CC).}
	\label{table:distractor}
\end{table}

% \subsection{Analysis of Functional Contrastive Learning}
% \subsection{Does FCL improve CNP?}

\begin{figure*}
     \centering
     \begin{subfigure}[b]{0.24\textwidth}
         \centering
         \includegraphics[width=\textwidth]{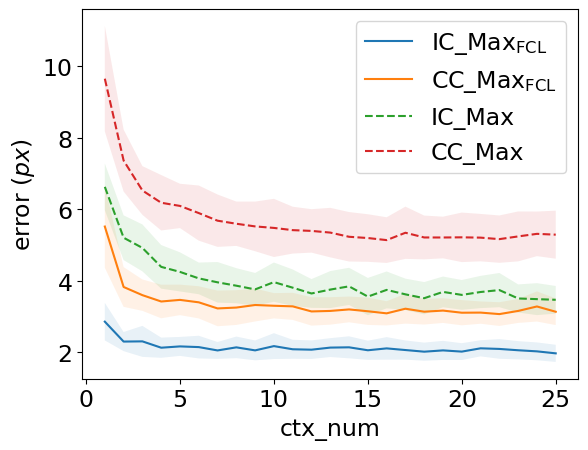}
         \caption{Distractor}
         \label{fig:loss_vs_ctx_distractor}
     \end{subfigure}
     \begin{subfigure}[b]{0.245\textwidth}
         \centering
         \includegraphics[width=\textwidth]{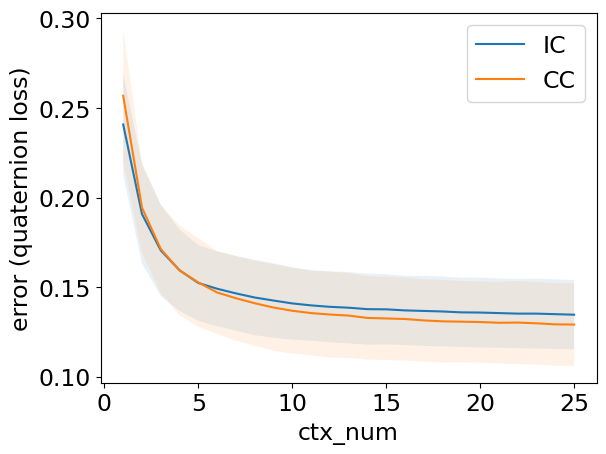}
         \caption{ShapeNet2D}
         \label{fig:loss_vs_ctx_shapenet3d}
     \end{subfigure} 
     \begin{subfigure}[b]{0.238\textwidth}
         \centering
         \includegraphics[width=\textwidth]{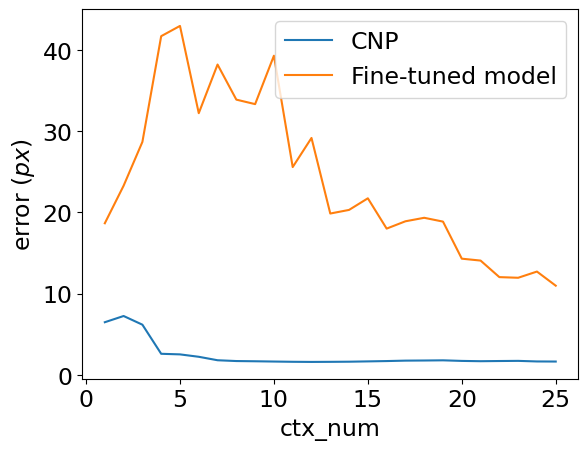}
         \caption{Distractor}
         \label{fig:finetune_distractor}
     \end{subfigure}
     \begin{subfigure}[b]{0.243\textwidth}
         \centering
         \includegraphics[width=\textwidth]{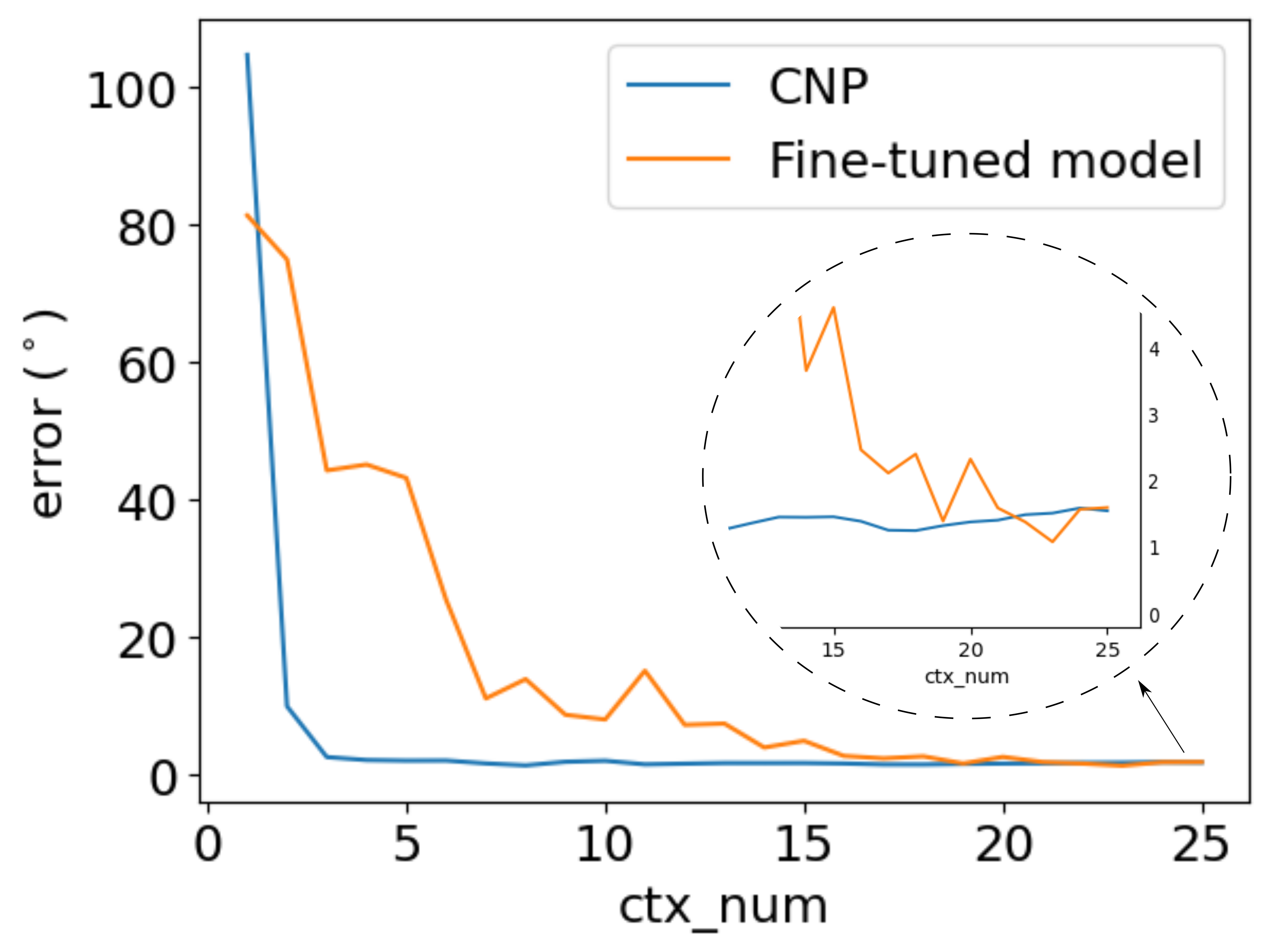}
         \caption{ShapeNet1D}
         \label{fig:finetune_shapenet1d}
     \end{subfigure}
    \caption{(a) CNP Prediction error (pixel) vs context number for the Distractor task using Max aggregation and Max + FCL (Max$_{\rm FCL}$). Results are evaluated on novel objects from both intra-category (IC) and unseen cross-category (CC) levels. (b) CNP (CA) Prediction error vs context number for ShapeNet2D using DA + TA. (c) We compare a classical object detection method and CNP (Max) using same dataset for training on Distractor. The classical model is further fine-tuned on each new task. The results are shown in dependence of the number of images used for fine-tuning or as context set.(d) Prediction error between the fine-tuned model and CNP (CA) on ShapeNet1D.}
    \label{fig:error_and_finetune}
\end{figure*}
\begin{table}[htb!]
	\centering
	\begin{tabular}{llll}
		\toprule
		Methods     & MAML             & CNP (Mean)                & CNP (CA)\\
		\midrule
		No Aug      & \round{1.6947} (\round{0.2219})  & \round{5.2759} (\round{0.5101})    & \round{4.6557} (\round{0.7445})\\
		MR          & \round{1.8993} (\round{0.2652})  & \round{2.9556} (\round{0.2097})    & \round{3.3307} (\round{0.2690})\\
		TA          & \textbf{\round{1.0204} (\round{0.0645})}  & \textbf{\round{1.9809} (\round{0.2183})}    & \textbf{\round{1.3638} (\round{0.2459})}\\
		DA          & \round{2.0967} (\round{0.0864})  & \round{3.6940} (\round{0.1260})    & \round{2.8957} (\round{0.0272})\\
	    TA+DA       & \round{1.3106} (\round{0.1400})  & \round{2.2903} (\round{0.1906})    & \round{1.7721} (\round{0.3316})\\
		\bottomrule
	\end{tabular}
	\caption{Pascal1D pose estimation error. MSE and standard deviations are calculated with 5 random seeds.}
	\label{table:pascal1d}
\end{table}
% \subsection{DA, TA or MR}
\begin{table}[t!]
	\centering
	\begin{tabular}{llll}
		\toprule
		Methods     & MAML             & CNP (Max)                & CNP (CA)        \\
		\midrule
		No Aug      & \round{25.2742}          & \round{14.9695} (\round{0.3671})    & \round{8.1941} (\round{0.3026})\\
		            & \round{21.633}           & \round{18.0917} (\round{0.2077})    & \round{9.1297} (\round{0.1768})\\
		MR          & \round{13.2298}          & \round{12.7079} (\round{0.2614})    & \round{8.8668} (\round{0.3638})\\
		            & \round{16.5522}          & \round{14.7687} (\round{0.3488})    & \round{8.4283} (\round{0.3904})\\
		TA          & \round{23.0069}          & \round{10.8864} (\round{0.2666})    & \round{7.9200} (\round{0.2537})\\
		            & \round{20.5857}          & \round{14.4286} (\round{0.5513})    & \round{9.1796} (\round{0.4974})\\
		DA          & \round{14.6929}          & \round{8.6409} (\round{0.2071})     & \round{6.2425} (\round{0.1544})\\
		            & \round{16.0171}          & \round{9.8736} (\round{0.3494})     & \round{6.5352} (\round{0.1858})\\
	    TA+DA       & \round{17.9595}          & \round{7.6582} (\round{0.1767})     & \textbf{\round{5.8051} (\round{0.2258})}\\
		            & \round{18.7901}          & \round{8.6554} (\round{0.1948})     & \textbf{\round{6.2267} (\round{0.1219})}\\
   		TA+DA+FCL   & {$\hspace*{7pt}-$}          & \round{7.8196} (\round{0.0811})      & \round{6.4374} (\round{0.3604})     \\
                    & {$\hspace*{7pt}-$}         & \round{8.8420} (\round{0.0413})      & \round{6.7437} (\round{0.1951})    \\
                    
		TA+DA+MR    & \round{13.4545}          & \round{10.5417} (\round{0.3716})    & \round{8.2810} (\round{0.1719})\\
	                & \round{14.4443}          & \round{10.7648} (\round{0.2992})    & \round{8.0441} (\round{0.1003})\\

		\bottomrule
	\end{tabular}
	\caption{ShapeNet1D pose estimation error($^\circ$). Results are calculated with 5 random seeds except for MAML. The first row presents results for IC and the second row for CC.}
	\label{table:shapenet1d}
\end{table}

\noindent\textbf{MAML or CNPs?}
We compare MAML and CNPs on two pose estimation datasets, Pascal1D and ShapeNet1D. We obtain similar results as \newcite{MWM, MRM} on Pascal1D, where MAML performs better than CNPs and the latter shows more severe overfitting (see \cref{table:pascal1d}).
However, \cref{table:shapenet1d} illustrates that both CNP variants outperform MAML with a large margin on ShapeNet1D.
It is good to note that, the prediction errors of all methods in \cref{table:pascal1d} after denormalizing are larger than 30$^\circ$, indicating the experiments of prior work on Pascal1D simply used too little meta-data to make informative conclusions about the quality of different algorithms.
Our interpretation is that  MAML tries to learn a good initial prior (global optimum) which needs to be optimized on each specific task (fine-tuned optimum) within few samples and updates. On small datasets, MAML can easily find a global optimum that satisfies all the training tasks. At the same time MAML also overfits less, since the fine-tuning from global to fine-tuned optimum happens during inference time.
However, finding a global optimum is getting difficult for large-scale datasets due to the increasing task diversity. Consequently, more samples and updates are necessary to fine-tune the task-specific parameters $\phi$ (see \cref{sec:background and notations}), which also explains why MAML is sensitive to hyperparameter tuning\newcite{howtotrainMAML}. Furthermore, MAML shows much longer training times than CNPs, which limits us to conduct exhaustive comparisons on more complicated tasks such as Distractor or ShapeNet2D. In contrast, CNPs use the local parameterization $\phi$ as a fixed dimensional output of the encoder, which forces the model to learn an informative low-rank representation from the contexts. Meanwhile, increasing data and task diversity will encourage the model to extract more expressive and mutual-exclusive task representations. 
% \comment{In addition, CNP explicitly disentangles the meta-parameters $\theta$ and task-specific parameter $\phi$\comment{ during meta-train and reduces the learnable parameters $\phi$ to a fixed number during meta-test compared to MAML} and thus reduces the number of parameters that are adaptable at test time.}
\\

\noindent\textbf{DA, DR, TA or MR?}
From the results of different experiments presented in  \cref{table:distractor}, \cref{table:pascal1d}, \cref{table:shapenet1d} and \cref{table:shapenet3d}, it is obvious that DA improves the performance across all tasks and methods. \cref{table:shapenet3d} also shows the importance of DR on ShapeNet2D which cannot be simply compensated by DA. TA hinders the performance on Distractor but benefits all pose regression tasks. The reason for this is that, for Distractor, TA increases task complexity by shifting the origin of the image plain by the sampled noise, thus creating $\rm{N}^2$ copies of the original task, where $\mathrm{N}=16$ is the number of non-zero elements in the noise set. However, since these task copies live in independent coordinate frames, the increased task diversity is irrelevant to the original task.
% \comment{from the original image plane coordinates $\mathbb{R}^2$ to $(\rm{N}^2+1)\times \mathbb{R}^2$ different coordiantes where N is the number of non-zero elements in the noise set, which is $16$ in our case (see \cref{exp:augmentation}). Specifically, the origin of the image plane is changing due to the sampled noise which accordingly varies the position of the queried object though the image remain the same. However, the increased task diversity (\ie various image plane coordinates) \comment{cannot benefit}\checklater{is irrelevant} to the original task and thus leads to severe underfitting. }
For pose regression tasks, by contrast, TA augments the canonical poses of the existing data, which coherently benefits the original task as the augmented canonical poses remain in the coordinate frame of the original task.
% \comment{For pose regression tasks, by contrast, \checklater{TA increases the canonical poses for the existing data and coherently benefits the original task since different canonical poses of different objects are inherent in the original task.}}
% \comment{by contrast, TA does not increase the existing coordinates either for 1D rotation $SO(1)$ or 2D rotation $SO(3)$. }
Therefore, even though TA increases the cross-entropy $\mathcal{H}(Y|X)$ for both cases as demanded in\newcite{MRM},
% (\ie different examples with shared input but different $y: (x, y+\epsilon_1), (x, y+\epsilon_2)$), 
only the pose regression tasks gain additional benefits. MR results in underfitting as combining MR with augmentations leads to worse performance than using the same augmentations alone for both ShapeNet1D and ShapeNet2D (see \cref{table:shapenet1d} and \cref{table:shapenet3d})\comment{, though using MR alone can alleviate overfitting (see \cref{table:pascal1d} and \cref{table:shapenet1d})}.
% MR alleviates overfitting without the use of other augmentations. However, combining MR with sophisticated augmentations can improve the performance in comparison to using solely augmentations (\cref{table:shapenet1d} and \cref{table:shapenet3d}). 
Furthermore, MR requires extensive fine-tuning on the regularization parameter $\beta$ (see \cref{exp:augmentation}) to modulate between underfitting and overfitting.\\

\begin{table}[t!]
	\centering
	\begin{tabular}{lll}
		\toprule
		Methods     & IC ($1e^{-2}$)                                   & CC ($1e^{-2}$)        \\
		\midrule
		None        & \round{38.33} (\round{0.33})                  & \round{39.81} (\round{0.31})    \\
		
		DR        & \round{18.67} (\round{0.13})              & \round{20.05} (\round{0.12})    \\
		
		DR+MR       & \round{27.89} (\round{0.61})                  & \round{28.99} (\round{0.46})    \\
		
	    DR+TA$_{\rm azi}$ & \round{16.94} (\round{0.13})                & \round{18.42} (\round{0.26})     \\
	    
	    DR+TA$_{\rm azi+ele}$ & \round{16.62} (\round{0.12})            & \round{17.76} (\round{0.35})     \\
		
		DA          & \round{19.32} (\round{0.09})                  & \round{17.98} (\round{0.09})    \\

		DR+DA       & \round{14.26} (\round{0.09})                  & \round{13.91} (\round{0.14})     \\
		
	    DR+DA+TA$_{\rm azi+ele}$  & {\round{14.12} (\round{0.14})}          & {\round{13.59} (\round{0.10})}     \\
	    
	    DR+DA+TA$_{\rm azi+ele}$ + FCL  & \textbf{\round{14.01} (\round{0.09})}          & \textbf{\round{13.32} (\round{0.18})}\\
		            
		\bottomrule
	\end{tabular}
	\caption{Comparison of different augmentation techniques on ShapeNet2D. Results are calculated with 3 random seeds using CNP (CA) as baseline.}
	\label{table:shapenet3d}
\end{table}

\begin{table*}
	\centering
	\begin{tabular}{lllllll}
		\toprule
		Methods     & CA$_{\rm S}$              & CA$_{\rm M}$                & CA$_{\rm L}$          & Max$_{\rm S}$              & Max$_{\rm M}$              & Max$_{\rm L}$\\
		\midrule
		No Aug      & \round{18.6016} (\round{0.7834})   & \round{12.0818} (\round{0.4353})   & \round{8.1941} (\round{0.3026})   & \round{30.4418} (\round{0.8177})   & \round{18.8604} (\round{0.3382})   & \round{14.9695} (\round{0.3671})   \\
		            & \round{19.9511} (\round{1.0800})   & \round{12.6172} (\round{0.8690})   & \round{9.1297} (\round{0.1768})   & \round{30.5896} (\round{1.1361})   & \round{21.7847} (\round{0.4652})   & \round{18.0917} (\round{0.2077})    \\
		TA          & \round{18.6932} (\round{0.8665})   & \round{10.7045} (\round{0.9812})   & \round{7.9200} (\round{0.2537})   & \round{21.6741} (\round{0.6576})   & \round{13.6857} (\round{0.2745})   & \round{10.8864} (\round{0.2666})    \\
		            & \round{19.2442} (\round{0.7914})   & \round{12.0488} (\round{0.7298})   & \round{9.1796} (\round{0.4974})   & \round{23.5958} (\round{0.8789})   & \round{16.7586} (\round{0.6195})   & \round{14.4286} (\round{0.5513})    \\
		TA+DA       & \round{7.8558} (\round{0.2074})    & \round{6.3165} (\round{0.1139})    & \textbf{\round{5.8051}} (\textbf{\round{0.2258}})   & \round{11.0000} (\round{0.1639})   & \round{8.2292} (\round{0.3363})    & \round{7.6582} (\round{0.1767})     \\
		            & \round{7.4945} (\round{0.3493})    & \round{6.4836} (\round{0.4112})    & \textbf{\round{6.2267}} (\textbf{\round{0.1219}})   & \round{12.9801} (\round{0.4822})   & \round{9.6538} (\round{0.3992})    & \round{8.6554} (\round{0.1948})     \\
		\bottomrule
	\end{tabular}
	\caption{Performance on ShapeNet1D using small ($\rm S$), medium ($\rm M$) and large ($\rm L$) training dataset sizes for CNP with cross-attention (CA) and Max aggregation. The first row presents results for intra-category (IC) and the second row for cross-category (CC) evaluation. MSE and standard deviations are calculated with 5 random seeds.}
	\label{table:shapenet1d_datasize}
\end{table*}

% \subsection{Effect of the context set size in CNP}
\noindent\textbf{Effect of the context set size in CNPs.}
We compare the prediction error \wrt the size of the context set for Distractor (see \cref{fig:loss_vs_ctx_distractor}) and ShapeNet2D (see \cref{fig:loss_vs_ctx_shapenet3d}). Both figures show that increasing the context set size benefits the performance, indicating that both Max and CA aggregations can merge useful information from different context pairs and thereby reduce the task ambiguity. In addition, we find that the model can further improve the performance given the size of context set surpasses the maximum number used for training (15 for both tasks). In particular, there is a small performance gap between intra- and cross-category evaluation for Distractor which is however absent for ShapeNet2D. We believe this indicates that Distractor has more task ambiguity than pose estimation and thus explains why Distractor gains more benefits from FCL than ShapeNet2D (see \cref{table:distractor} and \cref{table:shapenet3d}).
\\

% \subsection{CNP vs pretrained model}
\noindent\textbf{CNPs vs pretrained models.}
It is a common practice in vision task to pretrain a model\comment{Furthermore, pretrained model} on a large-scale dataset (\eg ImageNet\newcite{imagenet}) in order to obtain good prior features and reduce training time. To conduct a fair comparison of this approach to our model regarding data efficiency, we first pretrain a classical object detection model jointly over all tasks using the same training data as for CNPs. After training has finished, we fine-tune the pretrained model further on each specific new task using different numbers of images. Results are shown in \cref{fig:finetune_distractor} for Distractor and \cref{fig:finetune_shapenet1d} for ShapeNet1D, where the horizontal axis denotes the number of images used for fine-tuning or as contexts for CNPs, respectively. Both figures show that CNPs outperform the pretrained model especially for small numbers of contexts. In the Distractor task, CNP (Max) outperforms the fine-tuned model with a large margin after 25 context images are given. Note that CNPs are capable of transferring to various tasks simultaneously. In contrast, the pretrained model requires separate tuning on each given task, which results in a decreased performance on prior learning tasks.\\
% \subsection{Ablation study}
% \label{exp:ablationstudy}
\begin{table}[t!]
	\centering
	\begin{tabular}{lll}
		\toprule
		Methods       & IC ($1e^{-2}$)                                   & CC ($1e^{-2}$)        \\
		\midrule
		CNP+Mean      & \round{15.04} (\round{0.08})       & \round{15.45} (\round{0.13})    \\
		CNP+Max       & \round{14.20} (\round{0.06})       & \textbf{\round{13.56}} (\round{0.28})   \\
		CNP+BA      & \round{14.16} (\round{0.08})        & \textbf{\round{13.56}} (\round{0.18})    \\
		CNP+CA & \textbf{\round{14.12}} (\round{0.14})       & \round{13.59} (\round{0.10})     \\
		\bottomrule
	\end{tabular}
	\caption{Comparison of aggregation methods on ShapeNet2D using DR+DA+TA. Results are calculated with 3 random seeds.}
	\label{table:shapenet3d_aggregation}
\end{table}

% \subsection{Which aggregation methods should I use?}
\noindent\textbf{Which aggregation methods should I use?}
Cross-attention (CA) performs better than mean aggregation on Pascal1D (see \cref{table:pascal1d}) and Max on ShapeNet1D (see \cref{table:shapenet1d}), while it achieves a similar performance to Max aggregation and BA on ShapeNet2D (see \cref{table:shapenet3d_aggregation}). In contrast, mean aggregation used in the original CNP performs the worst on both Pascal1D and ShapeNet2D. Our interpretation is that Mean assigns the same importance to each context while the other aggregation operators can allocate different weights. \comment{In Max, the posterior entropy of the represented function is non-increasing given more contexts\newcite{uncertaintyNP}. }Max assigns a weight of one to a context and zero to all others for each dimension of the representation while BA assigns the weights predicted by another neural network. Meanwhile, CA assigns importance by comparing the similarity between context inputs $\{x_{C}^{i}\}_{i=1}^K$ and target input $x_T$ at the feature space. 
\comment{BA encodes the importance directly in the latent space with additional estimated uncertainty and applies the Bayes rule to update the posterior.}\comment{Therefore it helps to identify the task as long as more informative context is added}

Furthermore, we find that CA achieves competitive results on all pose estimation tasks but performs slightly worse than BA and Max on Distractor (see \cref{table:distractor}) though still better than mean aggregation. This indicates that CA helps in learning representations for object-centric images. Distractor, however, contains objects with random locations, requiring the model to disregard positional information. Methods like CA, which compare similarity between contexts and target over feature space, face inherent difficulties on Distractor. This is due to the fact that CNNs, owing to their translational equivariant nature, are prone to encode some positional information into the extracted image features.
Consequently, CA, which compares the similarity directly on this feature space, inevitably forces the model to focus on positional similarity, which leads to a suboptimal allocation of importance.
\\

% \subsection{How much data does CNP need?}
\noindent\textbf{How much meta-data is essential?}
We split the training data of ShapeNet1D into subsets of three different sizes, with 10 objects per category for the small dataset (S), 30 objects per category for the medium dataset (M) and 50 for the large dataset (L). Afterwards, we test the performance of CNP with Max aggregation and CA on each of them. The results in \cref{table:shapenet1d_datasize} show that Max overfits on the small dataset by simply memorizing all training tasks while CA works much better. Moreover, CA trained on small dataset achieves a comparable performance with Max on large dataset after using TA and DA, and even outperforms Max on the cross-category level. Thus, we conclude that using CA in combination with augmentation techniques can drastically alleviate the overfitting problem and therefore requires less meta-data on object-centric vision tasks than Max. In contrast, MAML performs much worse on ShapeNet1D (L) (see \cref{table:shapenet1d}) than CNPs and thus hardly profits from an increased dataset.
\\

% \subsection{Data augmentation}
\noindent\textbf{Data augmentation.}
\cref{table:shapenet3d_augmentation} shows the effect of each individual data augmentation technique (see \cref{exp:augmentation}) on ShapeNet2D. The first row contains results obtained with all techniques applied jointly. In the other rows, one of the techniques is removed respectively. We find that removing \textit{Affine} leads to the worst performance which indicates that object-centric pose regression tasks are more sensitive to scale. On the other hand, omitting \textit{CropAndPad} even leads to an performance increase.\\
\begin{table}[htb!]
	\centering
	\begin{tabular}{lll}
		\toprule
		Methods             & Val               & Test        \\
		\midrule
		All                 & {0.1417}          & {0.1410}    \\
		           
		w/o CropAndPad      & {0.1412}          & {0.1368}   \\
		           
		w/o Affine          & \textbf{0.1623}   & \textbf{0.1743}     \\
		           
		w/o Dropout         & {0.1452}          & {0.1445}      \\
		
		w/o Contrast         & {0.1482}         & {0.1406}      \\
		
		w/o Brightness         & {0.1454}       & {0.1380}      \\
		            
		w/o Blur         & {0.1426}             & {0.1422}      \\		            
		\bottomrule
	\end{tabular}
	\caption{Comparison of different data augmentation techniques on ShapeNet2D using CNP (CA) + DR as baseline.}
	\label{table:shapenet3d_augmentation}
\end{table}
\begin{table}[htb!]
	\centering
	\begin{tabular}{lll}
		\toprule
		Methods             & IC                                 & CC        \\
		\midrule
		Same Ctx         & \round{2.3036} (\round{0.0374})& \round{3.4628} (\round{0.0552})\\
		           
		Diff Ctx         & \round{2.1561} (\round{0.0511})& \round{3.2496} (\round{0.0474})\\
		           
		Ctx \& Target   & \textbf{\round{2.0039}} (\textbf{\round{0.0155}})      & \textbf{\round{3.0462}} (\textbf{\round{0.0753}})     \\
		            
		\bottomrule
	\end{tabular}
	\caption{Analysis of FCL + CNP on different choices of positive pairs using: i) the same context set with different augmentations (Same Ctx), ii) different context sets from the same task (Diff Ctx), iii) context and target sets (Ctx \& Target). Prediction error (pixel) is calculated with 3 random seeds. }
	\label{table:fcl_on_different_sets}
\end{table}

\noindent\textbf{Does FCL improve CNPs?}
\cref{table:distractor} shows the evaluation on Distractor using different aggregation methods where Max$_{\rm FCL}$ denotes Max aggregation with FCL. Modulating task representation by functional contrastive learning (FCL) alleviates meta overfitting across all augmentation levels and thus achieves a significant improvement in performance. \cref{fig:loss_vs_ctx_distractor} further compares the performance of Max and Max$_{\rm FCL}$ for different context set sizes, showing that our methods can differentiate the queried object and distractors well, even for very small context sets. Furthermore, we investigate the influence of FCL on the predicted task representations over all 12 categories using different clustering metrics, where the results show that FCL leads to a more dispersed latent distribution compared to the original CNPs, which can improve generalization capability to unseen tasks. T-SNE visualizations of the task representations along with the results of cluster metrics are provided in \cref{app:FCL}.\\

% \subsection{FCL on different sets}
\noindent\textbf{FCL on different sets.}
We compare FCL on three choices of positive pairs: i) We use the same context set but with different data augmentations. ii) We use different context sets sampled from the same task. iii) We use context and target sets from the same task. We test the performance on Distractor using Max aggregation and DA. For each choice, we run three experiments with different seeds and present the average performance in \cref{table:fcl_on_different_sets}. Compared to \cref{table:distractor}, all three choices consistently outperform CNP (Max) while using FCL on context and target sets achieves the best performance.\\
% In order to enforce the model learn object-centric representation, each image contains two different objects which located at random position and with random azimuth rotation. The rendering process of each object is followed by\newcite{VERSA}, namely, we choose the largest 12 object categories from ShapeNetCoreV2\newcite{ShapeNet} and randomly choose 50 objects from each category in order to keep the categorical balance. After then, we render 36 images around each object with every 10 degrees in azimuth. Lastly, for each image within the object, we randomly sample another image from another object belongs to different categories, and combine them together without overlapping or truncation. We show the generated example in Figure \checklater{add figure ref}. Note that the example is considered only as one task and we generate 12 categories each with 50 objects, i.e. 600 tasks in total. 10 categories are used during training, 80\% of the objects within these 10 categories are used for training and the remained 20\% as new tasks for intra-category validation. The other 2 categories are used for cross-category evaluation. \checklater{add data split here or create another section?}

% \subsection{Limitations}
\noindent\textbf{Limitations.}
Since meta-learning algorithms are data-driven methods, generalization depends on the diversity of training tasks. However, the augmentation methods used in our work are limited in their capability of creating new and diverse training tasks. Therefore, using a generative model to enrich training data could be one possibility to achieve higher diversity. Furthermore, concerning the class of NPs, we restricted ourselves to the deterministic CNPs in the experiments. We leave an in-depth exploration of stochastic NPs on vision tasks to future work.

\comment{Thus, it is essential to increase prior knowledge with more sophisticated augmentations. Augmentations used in our work are however restricted within existing knowledge.}
% \noindent\textbf{Scalability} Since meta-learning is a type of data-driven algorithm, the generalization depends on the diversity of training tasks. Thus, it is essential to increase prior knowledge with more sophisticated augmentations. Augmentations used in our work are restricted within existing knowledge, using generative model to enrich training data could be one solution.

% \noindent\textbf{Representation} The task representation in CNP is learned implicitly using different aggregation methods. However, this aggregation can be considered as a bottleneck and thus loses important information from context. We add contrustive loss after aggregation in order to disentangle different tasks in a category-agnostic way. Future work can investigate further on latent representation using more effective aggregation or compositional representation.

% \input{docs/7_discussion}
\section{Conclusion}
\label{conclusion}
In this paper, we investigate MAML and CNPs on several image-level regression tasks and analyze the importance of different choices in mitigating meta overfitting. Furthermore, we provide insights and practical recommendations of different algorithmic choices for CNPs with respect to various task settings. In addition, we combine CNPs with functional contrastive learning in task space and train in an end-to-end manner, which significantly improves the task expressivity of CNPs. We believe that our work can lay the basis for future work on designing and implementing meta-learning algorithms in image-based regression tasks.
% We design and conduct extensive experiments on image-based regression tasks using MAML and CNP and raise a new paradigm of object detection task which could potentially be used in target identificaion and tracking. 
% Afterwards, our work quantitatively and analytically compares the performance of different aggregation methods in CNP and the properties between CNP and MAML. Furthermore, we systematically evaluate augmentation techniques in order to have a better understanding of meta-learning overfitting. Subsequently, the comparison between CNP and classical supervised learning shows advantage of meta-learning with respect to data efficiency and generalization ability. 

%%%%%%%%% References
% \newpage
\clearpage
%{\small
%\bibliographystyle{ieee_fullname}
%\bibliography{docs/References}
%}

%%%%%%%%% Appendix
\newpage
\clearpage
\appendix
\setcounter{figure}{0}  
\setcounter{table}{0}

\section{Appendix} 
\label{appendix}

\subsection{Functional Contrastive Learning on CNPs}
\label{app:FCL}

\begin{table}[htb!]
	\centering
	\begin{tabular}{cccccc}
		\toprule
		$\tau$             & 1.0         &0.5            &0.2        &0.07           &0.007           \\
		\midrule
		IC            & {8.5550}    & {8.9810}      & {8.8551}  & \textbf{7.8196}       & {8.1409}          \\
		           
		CC           & {10.4660}    & {10.5135}    & {10.5604} & \textbf{8.8420}       &  {9.3846}     \\
		          
		\bottomrule
	\end{tabular}
	\caption{Results of the evaluation on ShapeNet1D using different temperature values in FCL.}
	\label{apptable:FCL_temperature_shapenet1d}
\end{table}

\begin{table}[htb!]
	\centering
	\begin{tabular}{cccccc}
		\toprule
		$\tau$             & 1.0         &0.5            &0.2        &0.07           &0.007           \\
		\midrule
		IC            & {0.1564}    & {0.174}      & {0.1962, }  & {0.1441}       & \textbf{0.1401}          \\
		           
		CC           & {0.1594}    & {0.1758}    & {0.2089} & {0.1390}       &  \textbf{0.1332}     \\
		          
		\bottomrule
	\end{tabular}
	\caption{Results of the evaluation on ShapeNet2D using different temperature values in FCL.}
	\label{apptable:FCL_temperature_shapenet3d}
\end{table}

\begin{table}[htb!]
	\centering
	\begin{tabular}{lll}
		\toprule
		Methods             & Max                                 & Max$_{\rm FCL}$\\
		\midrule
		ARI $\uparrow$      & \round{0.2079}                       & \round{0.2001}    \\
		           
		MI  $\uparrow$      & \round{1.1270}                       & \round{1.0320}  \\
		           
		SS  $\uparrow$      & \round{0.3140}                        & \round{0.1494}     \\
		           
		CHI $\uparrow$      & \round{118.7283}                        & \round{18.8978}    \\
		            
	    DBI $\downarrow$    & \round{1.0006}                       & \round{1.6529}    \\
		            
		\bottomrule
	\end{tabular}
	\caption{Analysis of latent task representation on Distractor between Max and Max$_{\rm FCL}$ using various clustering metrics.}
	\label{apptable:cluster_metrics}
\end{table}

A grid search on hyperparameter $\tau$ is very expensive especially on vision tasks. Therefore, we search only on a discrete set $\{0.007, 0.7, 0.2, 0.5, 1.0\}$ and find that $\tau=0.07$ shows the best performance on ShapeNet1D and $\tau=0.007$ on ShapeNet2D. The results are shown in \cref{apptable:FCL_temperature_shapenet1d} and \cref{apptable:FCL_temperature_shapenet3d}.

\begin{figure}[t!]
     \centering
     \begin{subfigure}[b]{0.3\textwidth}
         \centering
         \includegraphics[width=\textwidth]{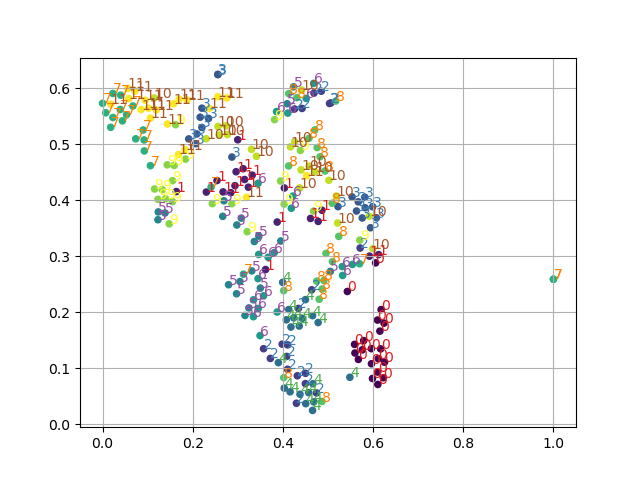}
         \caption{Max}
         \label{fig:finetune_distractor_max}
     \end{subfigure}\\
    %  \hfill
     \begin{subfigure}[b]{0.3\textwidth}
         \centering
         \includegraphics[width=\textwidth]{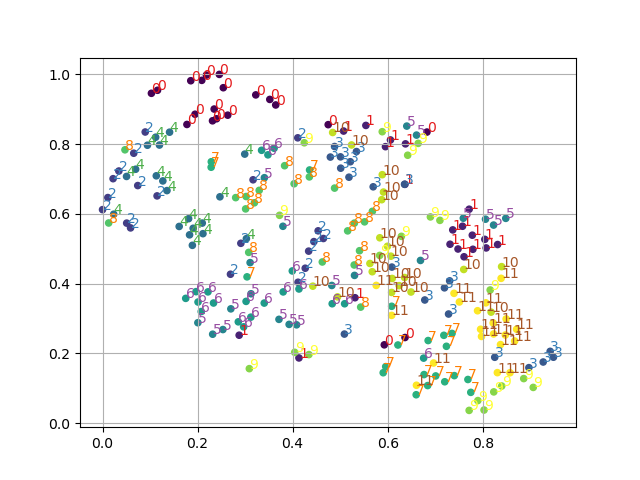}
         \caption{Max$_{\rm FCL}$}
         \label{fig:finetune_distractor_cmax}
     \end{subfigure}
    \caption{Visualization of latent variables on (a) max aggregation (b) max aggregation + functional contrastive learning (Max$_{\rm FCL}$).}
    \label{fig:tsne_distractor}
\end{figure}

For the Distractor task, we visualize the task representation obtained for novel objects in \cref{fig:tsne_distractor} where each color or number denotes one category and each point denotes the representation of each novel object. $\{10, 11\}$ are the novel categories \textit{\{sofa, watercraft\}}. Note that each object is considered as a single task and all tasks are learned in a category-agnostic manner. This figure indicates that Max$_{\rm FCL}$ can better shrink the distance between similar objects and repel the different ones implicitly. For instance, without a contrastive loss there is one outlier in \cref{fig:finetune_distractor_max} that is far away in representation space from the other objects. In particular, some samples are not well clustered based on categories, which is due to the high object variations within the same category.

Furthermore, we investigate the influence of FCL on the predicted task representations  over all 12 categories using five clustering metrics, namely Adjusted Rand Index (ARI), Mutual Information (MI), Silhouette Score (SS), Calinski-Harabasz Index (CHI) and Davies-Bouldin Index (DBI). Results are shown in \cref{apptable:cluster_metrics}. FCL leads to a more dispersed latent distribution compared to the original CNP, which reduces the vacancy in the latent space and thus improve the generalization ability to unseen tasks.

\subsection{Training Details}
For all tasks, we use 500k training iterations for CNPs and 70k for MAML. Furthermore, the best model on the intra- and cross-category dataset is saved during training. This leads to better models than early stopping with manually defined intervals. All experiments are conducted on a single NVIDIA V100-32GB GPU. Distractor and ShapeNet2D need around $3-5$ days for training, depending on different choices of augmentations, Pascal1D needs 8 hours and ShapeNet1D around 12 hours.

\textbf{Additional Results}. We have evaluated MMAML\newcite{MMAML}, a conditional variant of MAML, on ShapeNet1D based on reviewer's recommendation in \cref{apptable:MMAML_shapenet1d}. The results is worse than MAML, indicating that the designed task-aware modulation in MMAML doesn't benefit our tasks.

\begin{table}[htb!]
	\centering
	\begin{tabular}{ccccc}
		\toprule
		MMAML         & No Aug       &DA              &TA          &DA+TA                 \\
		\midrule
		IC            & {19.6900}    & {26.3624}      & {19.0705}     & {27.4973}          \\
		           
		CC           & {20.6123}    & {26.4090}    & {19.4285}      &  {27.3120}     \\
		\bottomrule
	\end{tabular}
	\caption{Performance of MMAML\newcite{MMAML} on ShapeNet1D.}
	\label{apptable:MMAML_shapenet1d}
\end{table}

\subsection{Task Augmentation}
The angular orientation of Pascal1D is normalized to $[0, 10]$ whereas ShapeNet1D uses radians with range $[0, 2\pi]$. For ShapeNet2D, the azimuth angles are restricted to the range $[0^\circ, 180^\circ]$ in order to reduce the effect of symmetric ambiguity while elevations are restricted to $[0^\circ, 30^\circ]$. we add random noise to both azimuth and elevation angles and then convert the rotation to quaternions for training.

\subsection{Data Augmentation}
\label{app:data_augmentation}
\textit{Affine} scales images between $80\%-120\%$ of their size along x and y axis and translate the images between $-10\%-10\%$ relative to the image height and width, and fills random value for the newly created pixels. \textit{Dropout} either drops random $1\%$-$10\%$ of all pixels or random image patches with $2\%-25\%$ of the original image size. \textit{CropAndPad} pads each side of the images less than $5\%$ of the image size using random value or the closest edge value. For ShapeNet2D, we furthermore add \textit{GammaContrast} with a range $[0.5, 2.0]$, \textit{AddToBrightness} with a range $[-30, 30]$ and \textit{AverageBlur} using a window of $k\times k$ neighbouring pixels where $k\in [1, 3]$. We use the open-source package\newcite{imgaug} for all data augmentations.

\begin{figure*}[ht]
     \centering
     \begin{subfigure}[b]{\textwidth}
         \centering
         \includegraphics[width=\textwidth]{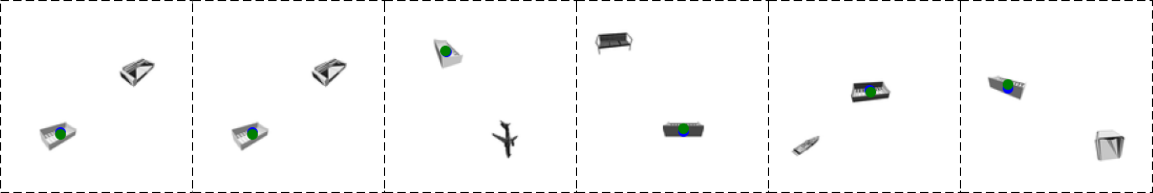}
         \caption{Distractor sofa}
         \label{appfig:example_distractor_sofa}
     \end{subfigure}\\
     \begin{subfigure}[b]{\textwidth}
         \centering
         \includegraphics[width=\textwidth]{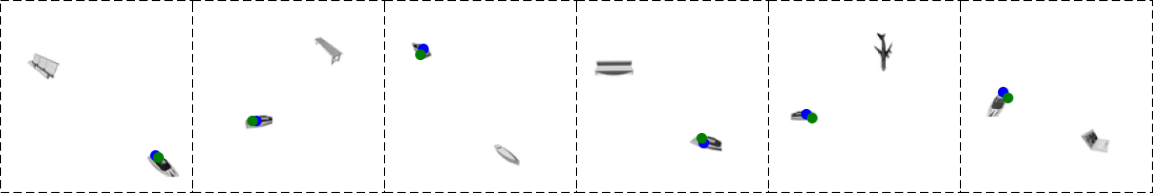}
         \caption{Distractor watercraft}
         \label{appfig:example_distractor_watercraft}
     \end{subfigure}\\
    \caption{Examples of Distractor on novel categories (sofa and watercraft) where green dots are ground-truth and blue dots are predicted positions.}
    \label{appfig:examples_distractor}
\end{figure*}
\begin{figure*}
     \centering
     \begin{subfigure}[b]{\textwidth}
         \centering
         \includegraphics[width=\textwidth]{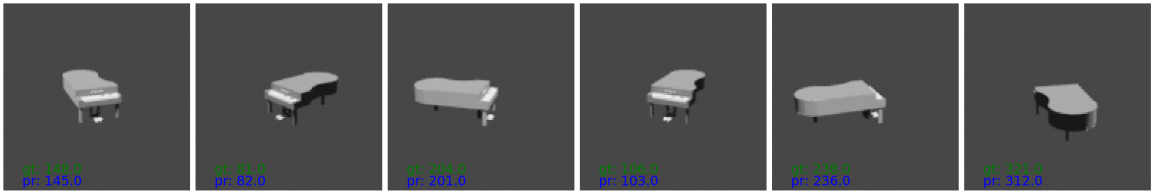}
         \caption{ShapeNet1D piano}
         \label{appfig:example_shapenet1d_piano}
     \end{subfigure}\\
    \begin{subfigure}[b]{\textwidth}
         \centering
         \includegraphics[width=\textwidth]{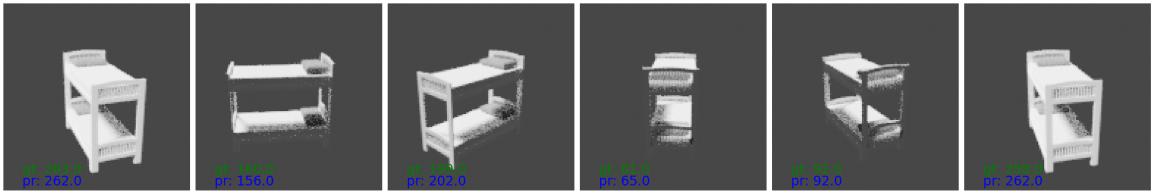}
         \caption{ShapeNet1D bed}
         \label{appfig:example_shapenet1d_bed}
    \end{subfigure}\\
    \begin{subfigure}[b]{\textwidth}
         \centering
         \includegraphics[width=\textwidth]{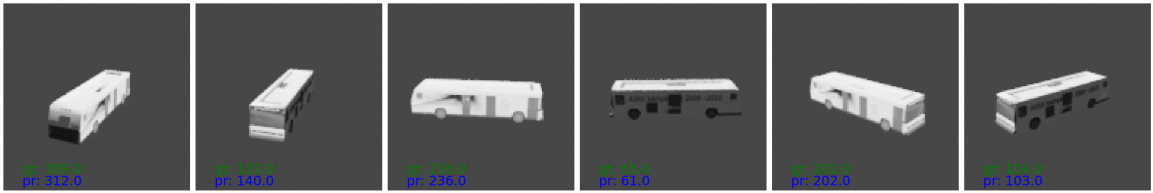}
         \caption{ShapeNet1D bus}
         \label{appfig:example_shapenet1d_bus}
    \end{subfigure}\\
    \caption{Examples of ShapeNet1D on novel categories (piano, bed, bus).}
    \label{appfig:examples_shapenet1d}
\end{figure*}

\subsection{Meta Regularization}
\label{app:meta-regularization}
Yin \etal\newcite{MWM} employ regularization on weights, the loss function is defined as:

\begin{align}
\label{appeq:loss_MR}
{\mathcal L} = \mathcal L_{O} + \beta {D_\mathrm{{KL}}}{(q(\theta;\theta_{\mu},\theta_{\sigma})||r(\theta))} 
\end{align}
where $L_{O}$ denotes the original loss function defined individually in Distractor and pose estimation. meta-parameters $\theta$ denote the parameters which are not used to adapt to the task training data. Function $r(\theta)$ is a variational approximation to the marginal which is set to ${\mathcal{N}(\theta;0,I)}$ in Yin \etal\newcite{MWM}. We follow the same setup in our experiments.
% \subsection{Network Architecture}
% \label{app:network_architecture}

\subsection{Examples of Inference Results}
\label{app:visualization_examples}

\begin{figure*}
     \centering
    \begin{subfigure}[b]{0.95\textwidth}
         \centering
         \includegraphics[width=\textwidth]{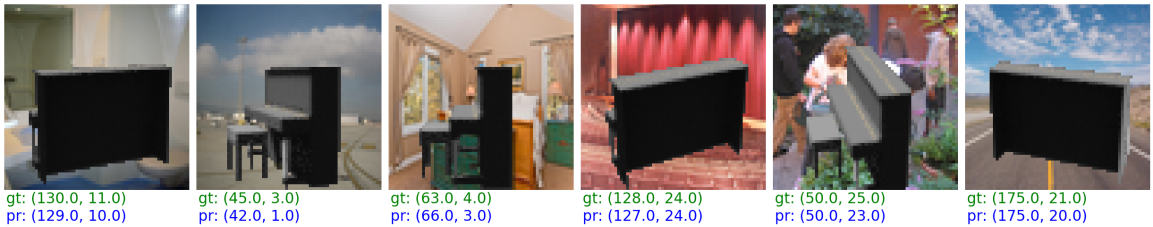}
        %  \caption{ShapeNet2D piano}
         \label{appfig:example_shapenet3d_piano}
    \end{subfigure}\\
    
    \begin{subfigure}[b]{0.95\textwidth}
         \centering
         \includegraphics[width=\textwidth]{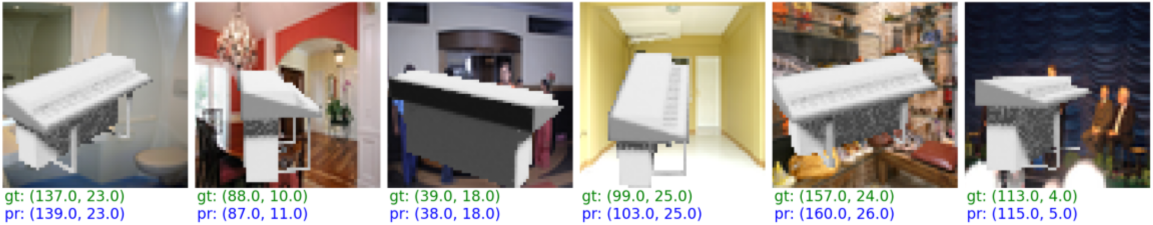}
        %  \caption{ShapeNet2D another piano}
         \label{appfig:example_shapenet3d_piano2}
    \end{subfigure}\\
    
    \begin{subfigure}[b]{0.95\textwidth}
         \centering
         \includegraphics[width=\textwidth]{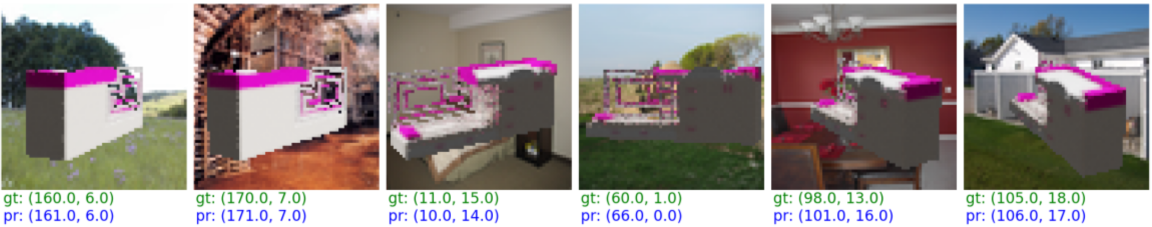}
        %  \caption{ShapeNet2D bed}
         \label{appfig:example_shapenet3d_bed}
    \end{subfigure}\\
    
    \begin{subfigure}[b]{0.95\textwidth}
         \centering
         \includegraphics[width=\textwidth]{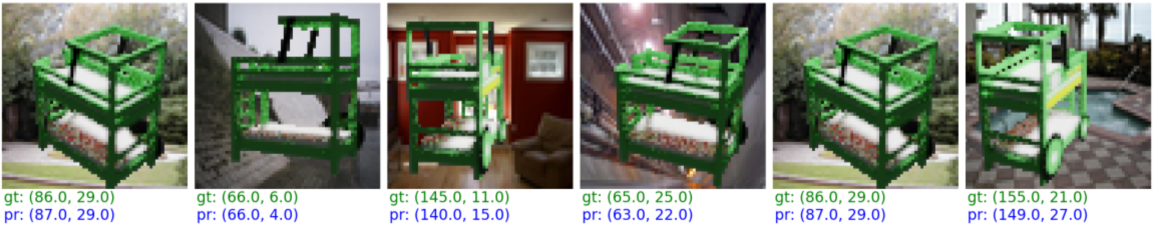}
        %  \caption{ShapeNet2D another bed}
         \label{appfig:example_shapenet3d_bed2}
    \end{subfigure}\\
    
    \begin{subfigure}[b]{0.95\textwidth}
         \centering
         \includegraphics[width=\textwidth]{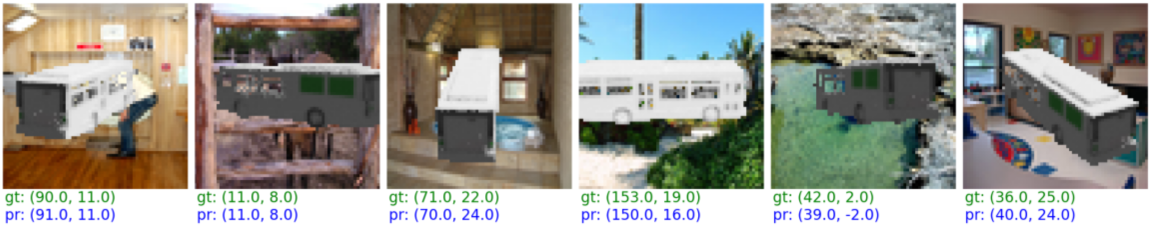}
        %  \caption{ShapeNet2D bus}
         \label{appfig:example_shapenet3d_bus}
    \end{subfigure}\\
    
    \begin{subfigure}[b]{0.95\textwidth}
         \centering
         \includegraphics[width=\textwidth]{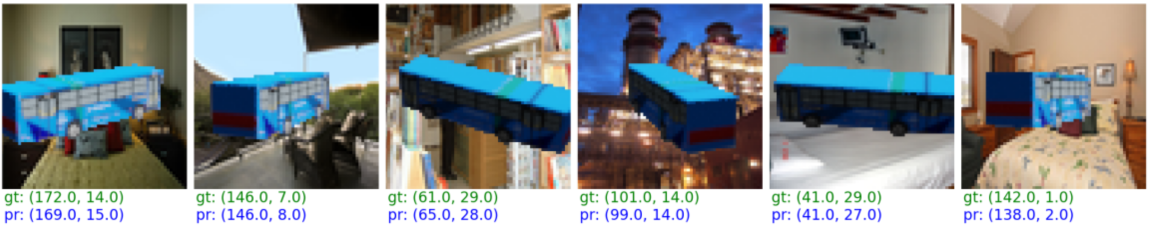}
        %  \caption{ShapeNet2D another bus}
         \label{appfig:example_shapenet3d_bus2}
    \end{subfigure}\\
    \caption{Examples of ShapeNet2D on novel categories (piano, bed, bus). Predictions are converted to (azimuth, elevation) angles.}
    \label{appfig:examples_shapenet3d}
\end{figure*}

We visualize examples of evaluation on novel cateogories in \cref{appfig:examples_distractor} for Distractor, \cref{appfig:examples_shapenet1d} for ShapeNet1D and \cref{appfig:examples_shapenet3d} for ShapeNet2D. 
% More results and code are available at \url{https://github.com/boschresearch/what-matters-for-meta-learning-vision-regression-tasks}.

% used for rebuttal link: \url{https://github.com/cvpr2022-5267/cvpr2022-id5267}

\end{document}